\documentclass[conference]{IEEEtran}
\IEEEoverridecommandlockouts
\usepackage{cite}
\usepackage{amsmath,amssymb,amsfonts}
\usepackage{algorithmic}
\usepackage{graphicx}
\usepackage{textcomp}
\usepackage{xcolor}
\def\BibTeX{{\rm B\kern-.05em{\sc i\kern-.025em b}\kern-.08em
    T\kern-.1667em\lower.7ex\hbox{E}\kern-.125emX}}
\begin{document}

\title{Building heterogeneous ensembles by pooling homogeneous ones\\
}

\author{\IEEEauthorblockN{Maryam Sabzevari}
\IEEEauthorblockA{\textit{Escuela Polit\'ecnica Superior} \\
	\textit{Universidad Aut\'onoma de Madrid}\\
	C/ Francisco Tom\'as y Valiente, 11 \\
	Madrid 28049, Spain\\
	maryam.sabzevari@uam.es}

\and
\IEEEauthorblockN{Gonzalo Mart\'{\i}nez-Mu\~noz}
\IEEEauthorblockA{\textit{Escuela Polit\'ecnica Superior} \\
	\textit{Universidad Aut\'onoma de Madrid}\\
	C/ Francisco Tom\'as y Valiente, 11 \\
	Madrid 28049, Spain\\
	gonzalo.martinez@uam.es}

\and
\IEEEauthorblockN{Alberto Su\'arez}
\IEEEauthorblockA{\textit{Escuela Polit\'ecnica Superior} \\
\textit{Universidad Aut\'onoma de Madrid}\\
	C/ Francisco Tom\'as y Valiente, 11 \\
	Madrid 28049, Spain\\
	alberto.suarez@uam.es}

}

\maketitle

\begin{abstract}
In ensemble methods, the outputs of a collection of diverse classifiers 
are combined in the expectation that the global prediction be more 
accurate than the individual ones. Heterogeneous ensembles consist of 
predictors of different types, which are likely to have different 
biases. If these biases are complementary, the combination of their 
decisions is beneficial. In this work, a family of heterogeneous 
ensembles is built by pooling classifiers from M homogeneous ensembles 
of different types of size T. Depending on the fraction of base 
classifiers of  each type, a particular heterogeneous combination in 
this family is represented by a point in a regular simplex in M 
dimensions. The M vertices of this simplex represent the different 
homogeneous ensembles. A displacement away from one of these vertices 
effects a smooth transformation of the corresponding homogeneous 
ensemble into a heterogeneous one. The optimal composition of such 
heterogeneous ensemble can be determined using cross-validation or, if 
bootstrap samples are used to build the individual classifiers, 
out-of-bag data. An empirical analysis of such combinations of 
bootstraped ensembles composed of neural networks, SVMs, and random 
trees (i.e. from a standard random forest) illustrates the gains that 
can be achieved by this heterogeneous ensemble creation method.
\end{abstract}

Note: This paper is under consideration at Pattern Recognition Letters.
\section{Introduction}
Building an effective classifier for a specific problem 
is a difficult task. To be successful,  
a variety of aspects need to be taken into account: 
the structure of the data, the information 
that can be used for prediction, the number of the
labeled examples available for induction, the noise level,
among others. Another crucial choice is the type of predictor
to be used. The strategies implemented
by the different classifiers are diverse. For instance, 
decision trees adopt a divide-and-conquer approach in 
which the original prediction task is recursively divided
by partitioning the attribute space  
into disjoint regions. Within each of these regions, 
the prediction problem is simpler than the original.
A neural network provides a global sub-symbolic representation 
of the decision problem in terms of the set of synaptic weights. 
Another illustration is the strategy adopted in kernel methods, 
such as Suppor Vector
Machines (SVM). In SVMs the original problem is  embedded
into an extended feature space. In this extended space, the 
discrimination problem is solved by finding the minimal margin
hyperplane that separates classes, except for, possibly, 
a few instances. In practice, one often finds that combining the outputs
of individual classifiers often leads to more accurate 
predictions. Whence, the popularity of ensemble methods
\cite{Dietterich_2000_Ensemble,Banfield_2000_Acomparison,Bauer_1999_An}. 
A necessary condition to obtain such improvements is that
the ensemble members be diverse. In additions, 
the individual predictors should be complementary,
in the sense that each of them tends to make errors on 
different test instances. 

Homogeneous ensembles are composed of  classifiers of the same type.
Ensembles composed of classifiers of different types are called
{\it{heterogeneous}}.
The strategies to generate diversity among the base classifiers 
are different for homogeneous and
for heterogeneous ensembles.
In homogeneous ensembles, the main difficulty is 
to generate diversity even when the same learning algorithm 
is used. To this end, one can use bootstrap techniques
(e.g. bagging \cite{breiman_1996_bagging}), randomized steps in the base learning 
algorithm (e.g. the random subspace method used random forest \cite{Breiman_2001_Random}),
noise injection in the class labels (e.g. class-swithcing \cite{martinezMunyoz+suarez_2005_switching}) 
or adaptive emphasis protocols (e.g. boosting \cite{freund++_1999_short}).
These techniques, which have exploited mainly in the context of homogeneous ensembles,
can also be used to achieve further diversity in heterogeneous
ensembles \cite{Lu_2015_Active}.
However, since different learning algorithms are used to 
generate the base learners,
heterogeneous ensembles are intrinsically diverse.
In this case, the main difficulty resides in determining the 
optimal way to combine the predictions of the different
models in the ensemble.

Broadly speaking, the methods to build heterogeneous ensemble
can be grouped into two categories. In the first family of methods 
a fixed number of different models are combined.
A second strategy is to build a collection of models with different 
parametrizations and then select the best subset to include in the final ensemble.
In \cite{Marques_2013_Ensemble} a static heterogeneous ensemble is proposed. In
this study 5 different base classifiers are combined: a Support Vector Machine (SVM), 
a multilayer perceptron (MLP), logistic regression, K nearest neighbors
and decision tree. The parameters and architecture of the individual
classifiers are determined using 10-fold cross-validation. The proposed approach shows
good results in the specific application of lithofacies classification. 
In \cite{nanni_2015_toward}, a combination of several carefully optimized strong
learners, such as deep neural networks, SVM, adaboosts, and gaussian processes, is
proposed. The study shows a good performance of the proposed combination over
several image classification and UCI tasks with respect to any of its
constituents. However, the problem of determining of 
the number of classifiers of each type that need to be used is not
solved in a fully satisfactory manner. 
Furthermore, the optimal composition of the ensemble is problem-dependent. 
A possible way to overcome this difficulty is to create a library of
classifiers and then select a subset for the final ensemble
\cite{Caruana_2004_Ensemble,Partalas_2010_An,haque_2016_heterogeneous}. 
For instance in \cite{Caruana_2004_Ensemble} a library of
2000 different methods trained with wide range of different parametrizations is
build. 
The models included in the library are both individual classifiers and ensembles. 
The ensemble methods used include boosted trees using different
decision tree algorithms and ensemble size, and bagged trees using different base
decision trees. In addition, the individual trees of the bagged ensembles were
also added to the library. Other individual classifiers included are SVMs
trained with different parameters, multilayer perceptrons, etc. From that
library of models, a iterative greedy selection algorithm is applied to build
the final ensemble. The procedure starts with empty ensemble. Then, at each
iteration the model that maximizes a performance measure (such as AUC
or accuracy on a validation set) is included into the ensemble until all
models in the library have been aggregated. Finally, the ensemble with the best
performance in the validation set is selected as the final combination.
Tsoumakas et al. have made several interesting contribution in this line of
research \cite{Tsoumakas_2004_Effective,Partalas_2010_An}. 
For instance, in \cite{Partalas_2010_An} the authors propose
a greedy selection method from a library composed of
200 classifiers: 60 neural networks, 60 nearest neighbor classifers,
80 SVMs and 20 decision trees). For each type of classifier, a parameter grid
was defined and a single model was trained for each node in the grid.
In their proposal, the ensemble is grown incrementally by selecting
from the library one classifier at a tiem. At each step, the 
selection is made in terms of both individual accuracy and 
complementarity with the rest of the classifiers in the ensemble. 
In the problems investigated, such heterogeneous ensembles were 
found to be more accurate that their constituents. 
In \cite{haque_2016_heterogeneous} a genetic algorithm has been proposed to
select the optimum structure of a heterogeneous ensemble from 20 different
base models. 
These selection techniques, also known as ensemble pruning, have been also
extensively applied to homogeneous ensembles \cite{Tsoumakas2009,martinezMunyoz++_2009_analysis}.

In this work we propose to analyze heterogeneous ensembles in 
which the individual classifiers are selected from homogeneous ensembles.
The goal is to build a family of heterogeneous ensembles
that can be smoothly transformed into each other another.
To this end, a family of heterogeneous ensembles of size T are built by 
pooling different fractions of base classifiers from M homogeneous ensembles 
of different types. Depending on the proportion of classifiers of each type, 
a particular heterogeneous combination in created.
This family of heterogeneous ensembles can be represented in a regular simplex 
in M dimensions. The M vertices of this simplex represent the different
homogeneous ensembles. 
The optimal fraction of each type of classifiers for the final ensemble
is found by performing a search is performed in this simplex.

The paper is organized as follows: Section \ref{Morph}, 
the design process to build optimal heterogeneous ensembles by pooling from
homogeneous ensembles is described; 
Section \ref{Exp}, presents a comprehensive empirical evaluation of 
the proposed methodology and a comparison with the
corresponding homogeneous ensembles and to individual
classifiers. Finally,the conclusions of the present work are summarized. 
\section{From homogeneous to heterogeneous ensembles} 
\label{Morph}
In this study we analyze in a systematic manner the construction of
heterogeneous ensembles by pooling individuals from different homogeneous
ensembles. For this, we first train $M$ ensembles of size $T$ composed of $M$
different types of base classifiers. The heterogeneous ensemble of size $T$ is
created by pooling $(t_1, t_2,\dots,t_M)$ classifiers from the $M$ ensembles,
where $t_j$ is the number of base classifiers pooled from the $j^{th}$
homogeneous ensemble and $\sum_{j=1}^M t_j = T$. The optimum percentage of each
type of base classifier can be obtained by cross-validation or out-of-bag error
in a grid search in the space given by $(t_1, t_2,\dots,t_M)$. Note, however,
that there are $\binom{T+M-1}{M-1}$ different heterogeneous ensembles that can be
built in this manner and that this number can be rather large even for small
values of $M$ and $T$. For instance, for $M=3$ and $T=101$, 5253 different
heterogeneous ensembles can be built. In order to reduce the search space, the
ensembles can be evaluated using intervals of $i$ base classifiers of each
type. For instance for $M=3$, the followings configurations of the generated
ensembles could be tested: $(0,0,T)$, $(0,i,T-i)$, $(i,0,T-i)$,
$(0,2*i,T-(2*i))$, $(2*i,0,T-(2*i))$, etc. This reduces the search space to
$\binom{T/i+M-1}{M-1}$ possible ensemble configurations. Finally, the ensemble
composition with minimum validation error is determined as the optimal ensemble.
In the case that more than one ensemble configuration has the same minimum
validation error, the average ensemble compositions for all minima with the same
validation error is selected as the optimal heterogeneous ensemble.

For this study, we have used three homogeneous ensembles: random forests (RF),
ensembles of support vector machines (SVMs) and of multilayer perceptrons
(MLPs). All base classifiers of these ensembles are created using random
samples from the training set to allow for a fast validation of the optimum 
heterogeneous ensemble by means of out-of-bag \cite{Breiman_1996_Out}. In order to generate 
ensembles of SVMs the following randomized procedure is used. First, $B$ sets of
partially optimized parameters for the SVMs, $\mathbf{\Theta}_b$ with $b=1,\dots,B$, are
obtained. More details on how these sets of partially optimized parameters are
obtained are given below. Then, the ensemble is built in $B$ batches of $T/B$
SVMs. Each batch uses a different set of parameters $\mathbf{\Theta}_b$ and each
individual SVMs is trained on a different random bootstrap sample without
replacement of size 50\% (i.e. subbagging) from the original training set. In
this way the variability among the SVMs can be increased. Using subbagging has the advantage
with respect to using standard bootstrap samples that the base models can be
trained faster. This speedup is approximately 4 times
considering the near quadratic training times of SVMs. In addition, the
performance of both sampling strategies, bootstrapping and subbagging, has been
demonstrated to be equivalent 
\cite{friedmann+hall_2007_bagging,martinezMunyoz+suarez_2010_out_of_bag}. 
To obtain the $B$ sets of partially optimized parameters, we first define a
parameter grid. Next, a subbagging sample is generated. One SVMs is trained for
each combination of parameters and validated on the left-out set. Finally, the
set of parameter with lower error is kept for building the ensemble. This
process is repeated $B$ times to obtain the $\mathbf{\Theta}_b$ with
$b=1,\dots,B$ sets of parameters. The same procedure is used to generate
the ensembles of MLPs. The training time complexity of the ensemble depends on
the size of the parameter grid, $B$, $T$, on the sampling rate and on the
complexity of the base classifier. In spite of creating an ensemble of SVMs (or
MLPs), this procedure can be faster to train than training a single SVM by grid
search and cross-validation, which is the most common way of training an SVM
\cite{Ben_2010_Auser,Hsu_2003_Apractical}. In the next section we will show the
validity of this procedure to generate homogeneous ensembles of SVMs and MLPs,
and also of the procedure to obtain heterogeneous ensembles from them.

\section{Experimental Results} \label{Exp}
In this section we present the empirical analysis of heterogeneous ensembles as
the combination of homogeneous base classifiers. Furthermore, we validate the
procedure to obtain SVM (and MLP) ensembles by partial optimization of their
training parameters. We carried out the analysis on
19 datasets from the UCI repository \cite{Bache+Lichman:2013}. In all tested
datasets, except of the synthetic problems, the training and test sets were
generated using random stratified sampling with sizes $2/3$ and $1/3$ of the 
original sets respectively. In the synthetic classification problems, which are
{\it Ringnorm}, {\it Threenorm} and {\it Twonorm}, $300$ examples are sampled at
random for training and $2000$ for testing using independent realizations. 
The results reported are averages over $100$ executions. 

Three, $M=3$, homogeneous ensembles of size $T=1001$ were trained. Specifically,
the ensembles used are: standard random forest \cite{Breiman_2001_Random}, partially
optimized ensemble of support vector machines \cite{Cortes_1995_Support} and of
multi layer perceptrons \cite{Haykin+1999}. We have used e1071, RSNNS and
randomForest R packages for creating SVMs, MLPs and RF respectively. Under these
setting the possible configurations of the heterogeneous ensemble are
$1003\times1002/2$. To reduce the computational burden to identify the optimum
combination of base classifiers, we evaluated the
heterogeneous ensembles in intervals of $i=13$ base learners, which reduces the optimization
to $78\times77/2$ evaluations. Given that all three ensembles were generated
using random subsamples from the training set to train each base classifier, the
optimum heterogeneous configuration is obtained by out-of-bag validation.
The values of the hyperparameters for SVM with a RBF kernel are selected from a
grid with $C=2^q$ with $q =-5, \dots, 15$ and $\boldsymbol{\gamma}=2^p$ with $p
=-15,\dots,3$. For MLP, the number of neurons in the hidden layer was optimized 
from the values $\{3,4,5,6,7,8,9,10\}$. For building the partially optimized
ensemble, $B=10$ sets of hyperparameter were obtained using out-of-bag. For
random forest, the default parameters were used.

\begin{table}
\footnotesize
\centering
\caption{Test errors for a single optimized SVM and MLP, also their
	homogeneous ensembles as it is proposed in section \ref{build_homo}}
\label{tab:errors_single_vs_ens} \centering
\resizebox{\columnwidth}{!}{%
	\begin{tabular}{@{}l@{$\;$}|
			r@{$\pm$}l@{$\;$}|r@{$\pm$}l@{$\;$}|r@{$\pm$}l@{$\;$}|r@{$\pm$}l@{$\;$}}
		\hline
		Dataset  &
		\multicolumn{2}{c|}{SVM} &
		\multicolumn{2}{c|}{E-SVM} &
		\multicolumn{2}{c|}{MLP} &
		\multicolumn{2}{c}{E-MLP} \\
		\hline
		Australian
		& 14.4 & 2.3 & 13.7 & 2.1 & 15.6 & 2.2 & 14.2 & 1.9\\
		\hline
		Boston
		& 12.2 & 2.4 & 12.2 & 2.3 & 12.7 & 2.1 & 12.3 & 2.0\\
		\hline
		Breast
		& 3.5 & 1.1 & 3.4 & 1.1 & 9.1 & 12.1 & 3.2 & 1.1\\
		\hline
		Bupa
		& 29.1 & 3.7 & 27.9 & 3.4 & 30.3 & 4.0 & 28.3 & 3.7\\
		\hline
		Chess
		& 0.8 & 0.4 & 0.8 & 0.3 & 1.0 & 0.2 & 0.9 & 0.3\\
		\hline
		Colic
		& 31.8 & 3.4 & 33.2 & 1.4 & 32.4 & 3.4 & 31.3 & 3.3\\
		\hline
		German
		& 25.1 & 1.8 & 24.6 & 1.6 & 28.0 & 2.0 & 24.7 & 1.9\\
		\hline
		Heart
		& 16.0 & 3.5 & 15.4 & 3.0 & 18.2 & 3.7 & 16.3 & 3.1\\
		\hline
		Hepatitis
		& 16.6 & 3.6 & 15.8 & 3.0 & 17.5 & 4.3 & 15.4 & 4.3\\
		\hline
		Ionosphere
		& 6.3 & 1.8 & 5.7 & 1.7 & 10.6 & 2.9 & 11.3 & 2.5\\
		\hline
		Ozone
		& 5.6 & 0.4 & 5.6 & 0.3 & 6.8 & 0.5 & 5.5 & 0.5\\
		\hline
		Parkinsons
		& 8.7 & 4.1 & 10.7 & 3.7 & 11.3 & 4.1 & 13.7 & 3.9\\
		\hline
		Pima
		& 23.1 & 2.0 & 22.7 & 1.8 & 24.7 & 2.4 & 23.1 & 2.1\\
		\hline
		Ringnorm
		& 1.7 & 0.6 & 1.6 & 0.4 & 17.0 & 1.5 & 16.4 & 1.5\\
		\hline
		Spambase
		& 6.4 & 0.4 & 6.6 & 0.4 & 7.0 & 0.4 & 5.9 & 0.4\\
		\hline
		Sonar
		& 15.0 & 4.3 & 17.8 & 4.9 & 21.0 & 4.2 & 20.7 & 4.6\\
		\hline
		Threenorm
		& 14.5 & 1.3 & 14.1 & 0.7 & 17.7 & 2.0 & 16.9 & 0.9\\
		\hline
		Tictactoe
		& 1.0 & 1.3 & 1.8 & 0.7 & 4.4 & 7.4 & 1.8 & 0.7\\
		\hline
		Twonorm
		& 2.6 & 0.5 & 2.4 & 0.3 & 22.4 & 22.9 & 2.9 & 0.4\\
	\end{tabular}}
\end{table}

\begin{figure}[tb]
	\centering
	\includegraphics[scale=0.5]{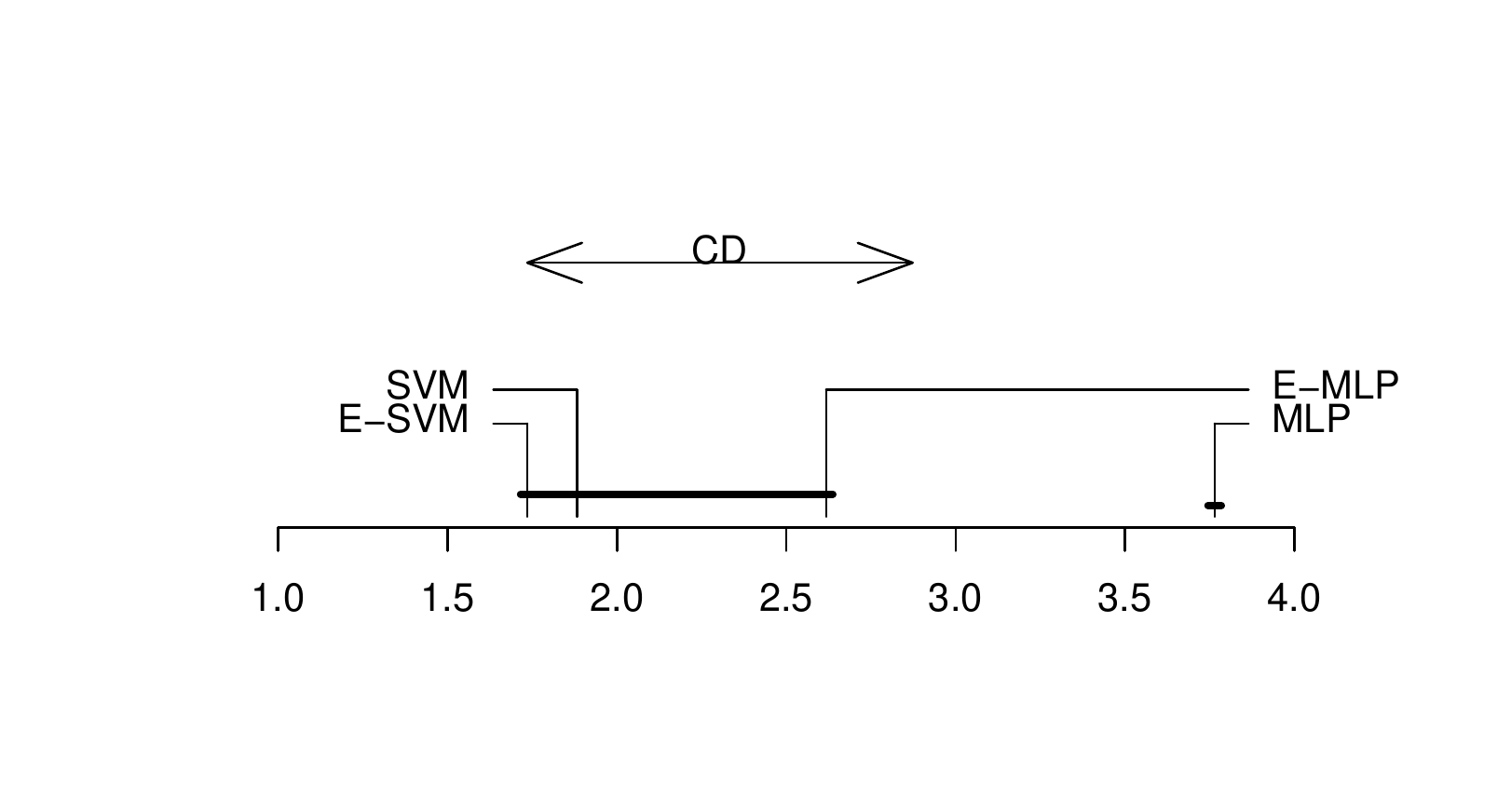}
	\caption{Average ranks for SVM, E-SVM, MLP and E-MLP (more details in the text)  
	}\label{demsar_sing_vs_ens}
\end{figure}

\subsection{Homogenous ensemble of SVMs and MLPs} \label{build_homo}
In order to validate the procedure to generate the partially optimized
ensembles, a comprehensive comparison with respect to an optimized single base
learner was carried out. For this purpose, a single SVM and a single MLP were 
trained using within-train 10-fold cross-validation and grid search over the 
same sets of parameters given above. The average errors for this experiments are
shown in Table~\ref{tab:errors_single_vs_ens} for a single SVM and MLP, and for
the homogeneous ensembles composed of SVMs (shown as E-SVM in the table) and of MLPs 
(shown as E-MLP).
In addition, 
an overall comparison of the methods is shown in Fig.~\ref{demsar_sing_vs_ens} 
by mean of the procedure proposed by Dem\v{s}ar in \cite{demsar_2006_statistical}.
In this diagram, the average ranks for each method are shown.
Methods connected by a horizontal solid line indicate that their differences in
average rank are not statistically significant according to a Nemenyi test
(p-value $<$ 0.05). 

From Table~\ref{tab:errors_single_vs_ens}, 
it can be observed that the ensemble of MLPs clearly outperforms the single MLP.
The ensemble of MLPs outperforms a single MLP in all tested datasets except for
\emph{Ionosphere} and \emph{Parkinsons}. The differences between the single SVM
and its ensemble counterpart are not so pronounced as the ones observed for
MLPs. The ensemble of SVM obtains a better result than a single SVM in 11 out of
19 datasets. This same result can be observed in Fig.~\ref{demsar_sing_vs_ens}
where the average rank of E-SVM is slightly better than a single SVM.
However, the difference is not statistically significant. 
Even thought the differences are not statistically significant, this analysis shows that this procedure to build
ensembles of SVMs is not detrimental. When using MLP as base
classifiers, we observe that the differences are statistically significant with
respect to a single MLP. 
In addition, with these setting, we have observed that the training time for
E-SVM is over 2 times faster than training a single SVM using grid search and
10-fold cross-validation. For ensembles of MLP, the speedup is over 1.5 with
respect to the single MLP.

\begin{figure*}[t]
	\centering
	\begin{tabular}{@{}c@{}c@{}c@{}}
		\includegraphics[width=0.33\textwidth]{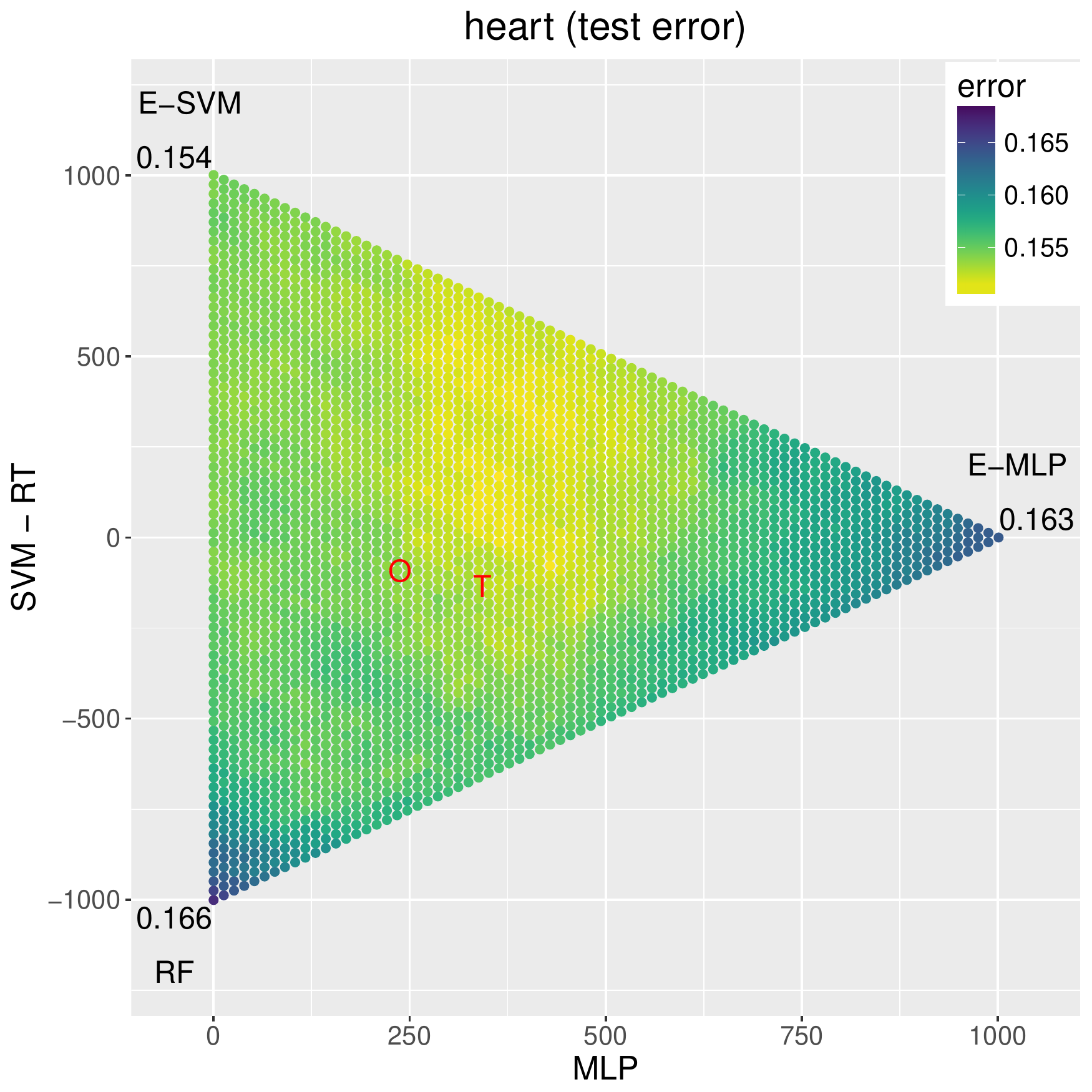}&
		\includegraphics[width=0.33\textwidth]{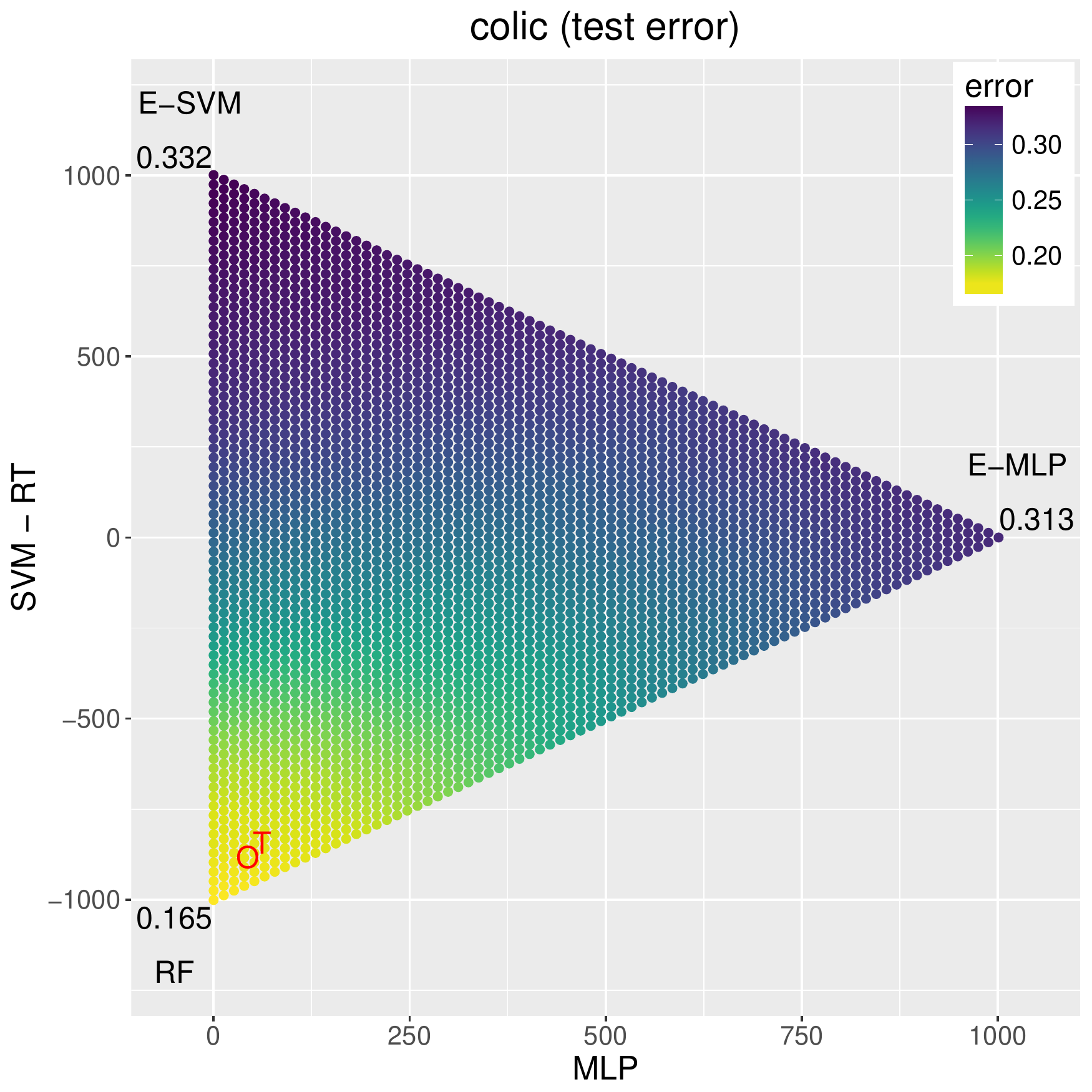}&
		\includegraphics[width=0.33\textwidth]{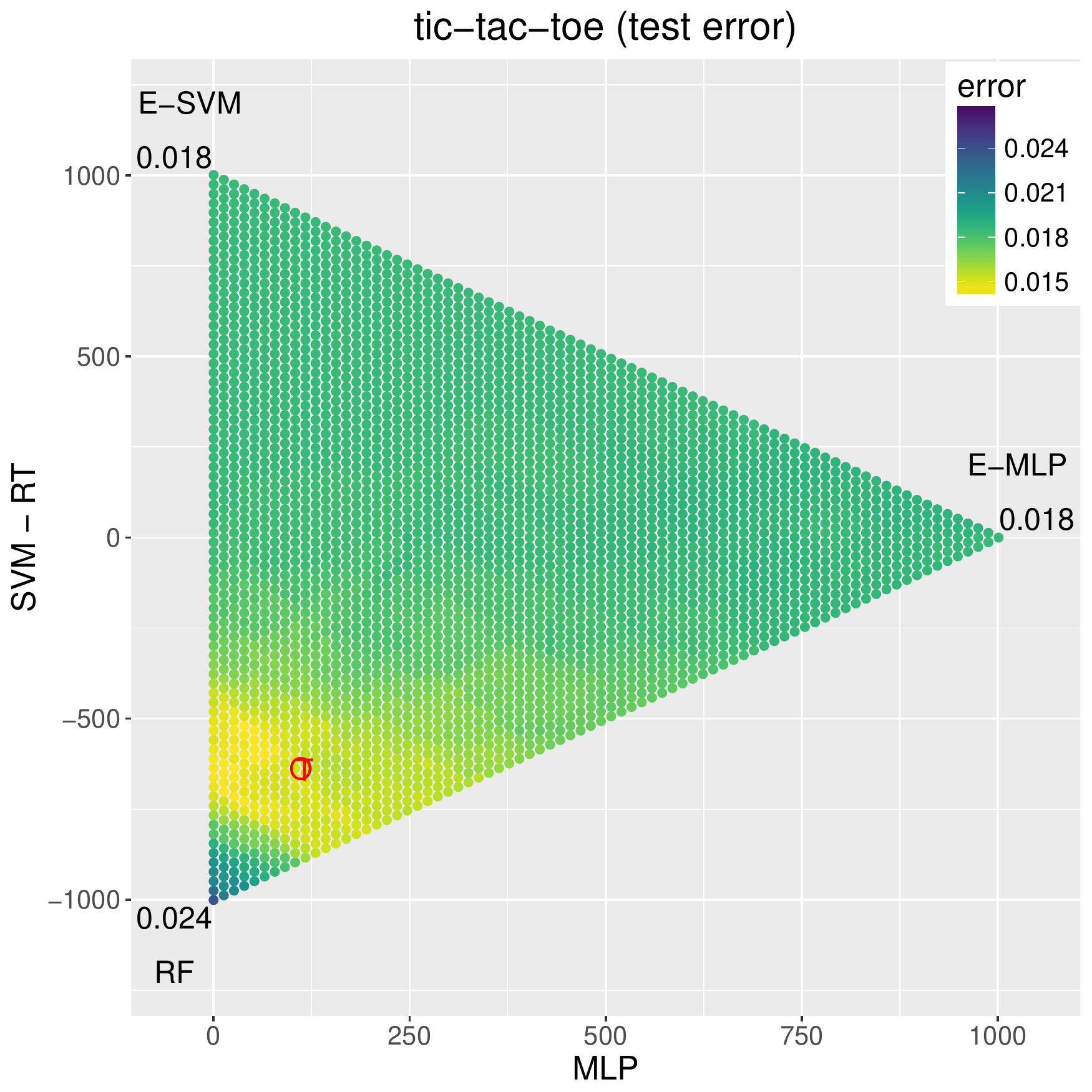}\\
	\end{tabular}
	\caption {Test error rate of the heterogenous ensembles in the simplex 
		for different classification problems. Darker colors correspond to 
		higher errors}
	\label{color-code-datasets}
\end{figure*}

\subsection{Heterogeneous ensemble pooled from homogeneous ensembles} \label{build_heter}
In this section the performance of the proposed procedure to built heterogeneous
ensembles by pooling from homogeneous ensembles is analyzed. The objective is to
find the optimum proportion of each of the possible base classifiers to build
the final heterogeneous ensemble. Each of the possible selected proportions,
which correspond to a different heterogeneous ensemble, can be represented by a 
point in a regular simplex in M dimensions. This is shown in
Fig.~\ref{color-code-datasets} for three representative datasets: {\it Heart},
{\it Colic} and {\it Tic-tac-toe}. Each plot in Fig.~\ref{color-code-datasets}
shows in a 3 dimensional simplex, the average test error for the different
combinations of base classifiers in intervals of $i=13$ classifiers using a grey
scale scheme. Darker colors indicate higher average error as indicated by the
color legend at the right of each plot. The three vertices in the plots
correspond to the three tested homogeneous ensembles. The vertices in the upper
left, right and bottom left of the plot correspond to E-SVM, E-MLP and random
forest respectively. A displacement away from one of these vertices smoothly
transforms the corresponding homogeneous ensembles into a heterogeneous one. The
horizontal axis shows the number of selected MLPs in the heterogeneous ensemble,
while the vertical axis indicates the number of SVMs minus the number of random
trees. In addition, all plots show the average selected position using
out-of-bag validation (marked with a 'o' sign) and the average position for
the best test errors (marked with a 'T' sign).

In the plots of Fig.~\ref{color-code-datasets} different behaviours of the
combination of base classifiers can be observed. In {\it Heart} (left plot), the
best position is observed quite centered, showing that a heterogeneous ensemble
composed of base classifiers from different types is beneficial to improve the 
generalization performance of the ensemble. However, this is not a general
trend as it can be observed in the center plot ({\it Colic}). In this case, the 
best result is clearly located at one of the vertices of the simplex that
correspond to a homogeneous ensemble ---random forest in this case. Finally, it
is important to note that the optimum location need not be close to the best
homogeneous ensemble. For instance, in {\it Tic-tac-toe}, the location of the
minimum error is very close to the random forest vertex in spite of the fact
that this homogeneous ensemble presents the worst average performance. Finally,
we can observe that the average location of the minima identified using
out-of-bag is quite close to the location in test. We have also observed,
however, that for the smaller datasets the identification of the optimum point
is less accurate. 

\begin{table*}
	\footnotesize
	\centering
	\caption{Test errors of single classifiers, homogeneous ensembles and
		optimal heterogeneous ensemble (I)}
	\label{tab:errors} \centering
	\begin{tabular}{@{}l@{$\;$}|
			r@{$\pm$}l@{$\;$}|r@{$\pm$}l@{$\;$}|r@{$\pm$}l@{$\;$}|r@{$\pm$}l@{$\;$}|c|c}
		\hline
		Dataset  &
		\multicolumn{2}{c|}{E-SVM} &
		\multicolumn{2}{c|}{E-MLP} &
		\multicolumn{2}{c|}{RF} &
		\multicolumn{2}{c|}{SIM} &[\% SVM, \% MLP, \% Trees]& entropy  \\
		\hline
		Australian
		& 13.7 & 2.1 & 14.2 & 1.9& {\bf 13.0} & {\bf 2.1}*& \underline{13.5} &
		\underline{2.0}&  [ 24.5 , 16.7 , 58.8 ] & 1.38 \\
		\hline
		Boston
		& {\bf 12.2} & {\bf 2.3} & 12.3 & 2.0 & 12.9 & 2.1& \underline{12.2} &
		\underline{2.0}&  [ 39.2 , 23.1 , 37.7 ] & 1.55 \\
		\hline
		Breast
		& 3.4 & 1.1& {\bf 3.2} & {\bf 1.1}& \underline{3.3} & \underline{1.1}
		& 3.3 & 1.0&  [ 27.6 , 29.0 , 43.4 ] & 1.55 \\
		\hline
		Bupa
		& 27.9 & 3.4 & 28.3 & 3.7& {\bf 27.2} & {\bf 3.6}& \underline{27.3} &
		\underline{3.5}&  [ 21.4 , 15.6 , 63.0 ] & 0.86 \\
		\hline
		Chess
		& \underline{0.8} & \underline{0.3} & 0.9 & 0.3 & 1.7 & 0.4& {\bf 0.8}
		& {\bf 0.2}&  [ 34.7 , 22.7 , 42.6 ] & 1.54 \\
		\hline
		Colic
		& 33.2 & 1.4 & 31.3 & 3.3& {\bf 16.5} & {\bf 2.9}*& \underline{17.2} &
		\underline{3.0}&  [ ~3.7 , ~4.4 , 91.9 ] & 0.48 \\
		\hline
		German
		& 24.6 & 1.6 & 24.7 & 1.9& {\bf 23.9} & {\bf 1.8}& \underline{24.3} &
		\underline{1.9}&  [ 15.9 , 28.3 , 55.7 ] & 1.41 \\
		\hline
		Heart
		& {\bf 15.4} & {\bf 3.0} & 16.3 & 3.1 & 16.6 & 2.9& \underline{15.5} &
		\underline{3.1}&  [ 33.5 , 23.8 , 42.7 ] & 1.55 \\
		\hline
		Hepatitis
		& 15.8 & 3.0 & 15.4 & 4.3& {\bf 15.1} & {\bf 3.6}& \underline{15.2} &
		\underline{3.6}&  [ 25.6 , 28.7 , 45.7 ] & 1.54 \\
		\hline
		Ionosphere
		& {\bf 5.7} & {\bf 1.7} & 11.3 & 2.5 & 6.7 & 1.7& \underline{5.8} &
		\underline{1.7}&  [ 64.4 , 13.7 , 21.9 ] & 1.28 \\
		\hline
		Ozone
		& 5.6 & 0.3& \underline{5.5} & \underline{0.5} & 5.7 & 0.3& {\bf 5.4}
		& {\bf 0.4}&  [ 16.7 , 53.6 , 29.7 ] & 1.43 \\
		\hline
		Parkinsons
		& {\bf 10.7} & {\bf 3.7} & 13.7 & 3.9& \underline{11.1} &
		\underline{4.0}& {\bf 10.7} & {\bf 3.9}&  [ 44.1 , 12.9 , 43.0 ] & 1.43
		\\
		\hline
		Pima
		& {\bf 22.7} & {\bf 1.8}* & 23.1 & 2.1 & 23.1 & 2.0& \underline{22.9}
		& \underline{1.8}&  [ 44.5 , 18.1 , 37.4 ] & 1.5 \\
		\hline
		Ringnorm
		& {\bf 1.6} & {\bf 0.4}* & 16.4 & 1.5 & 5.9 & 1.0& \underline{1.7} &
		\underline{0.5}&  [ 62.2 , 11.2 , 26.6 ] & 1.29 \\
		\hline
		Spambase
		& 6.6 & 0.4 & 5.9 & 0.4& \underline{5.1} & \underline{0.4}& {\bf 5.0}
		& {\bf 0.3}&  [ 12.2 , 11.1 , 76.8 ] & 1.01 \\
		\hline
		Sonar
		& {\bf 17.8} & {\bf 4.9} & 20.7 & 4.6 & 18.9 & 4.8& \underline{18.0} &
		\underline{4.4}&  [ 39.1 , 16.9 , 44.0 ] & 1.48 \\
		\hline
		Threenorm
		& {\bf 14.1} & {\bf 0.7}* & 16.9 & 0.9 & 16.7 & 1.0& \underline{14.4}
		& \underline{0.8}&  [ 52.1 , 10.8 , 37.1 ] & 1.37 \\
		\hline
		Tictactoe
		& \underline{1.8} & \underline{0.7}& \underline{1.8} & \underline{0.7}
		& 2.4 & 1.1& {\bf 1.5} & {\bf 0.7}*&  [ 12.5 , 11.2 , 76.3 ] & 1.02 \\
		\hline
		Twonorm
		& {\bf 2.4} & {\bf 0.3}* & 2.9 & 0.4 & 3.9 & 0.5& \underline{2.5} &
		\underline{0.4}&  [ 42.5 , 22.7 , 34.8 ] & 1.54 \\
		\hline
	\end{tabular}
\end{table*}

In the Table~\ref{tab:errors}, the average test errors for the homogeneous
ensembles of SVMs (E-SVM) and MLPs (E-MLP), random forest (RF) and the proposed 
strategy (SIM) over the investigated  problems are reported. The best and second
best results for each dataset are highlighted in boldface and underlined
respectively. In addition, the table shows the average percentage of classifiers
of each type selected by out-of-bag validation for the heterogeneous ensembles.
The percentages are shown in the same order that the ensembles are shown, that
is, \% of SVMs, \% of MLP and \% of random trees. 

\begin{figure}[tb]
	\centering
	\includegraphics[scale=0.5]{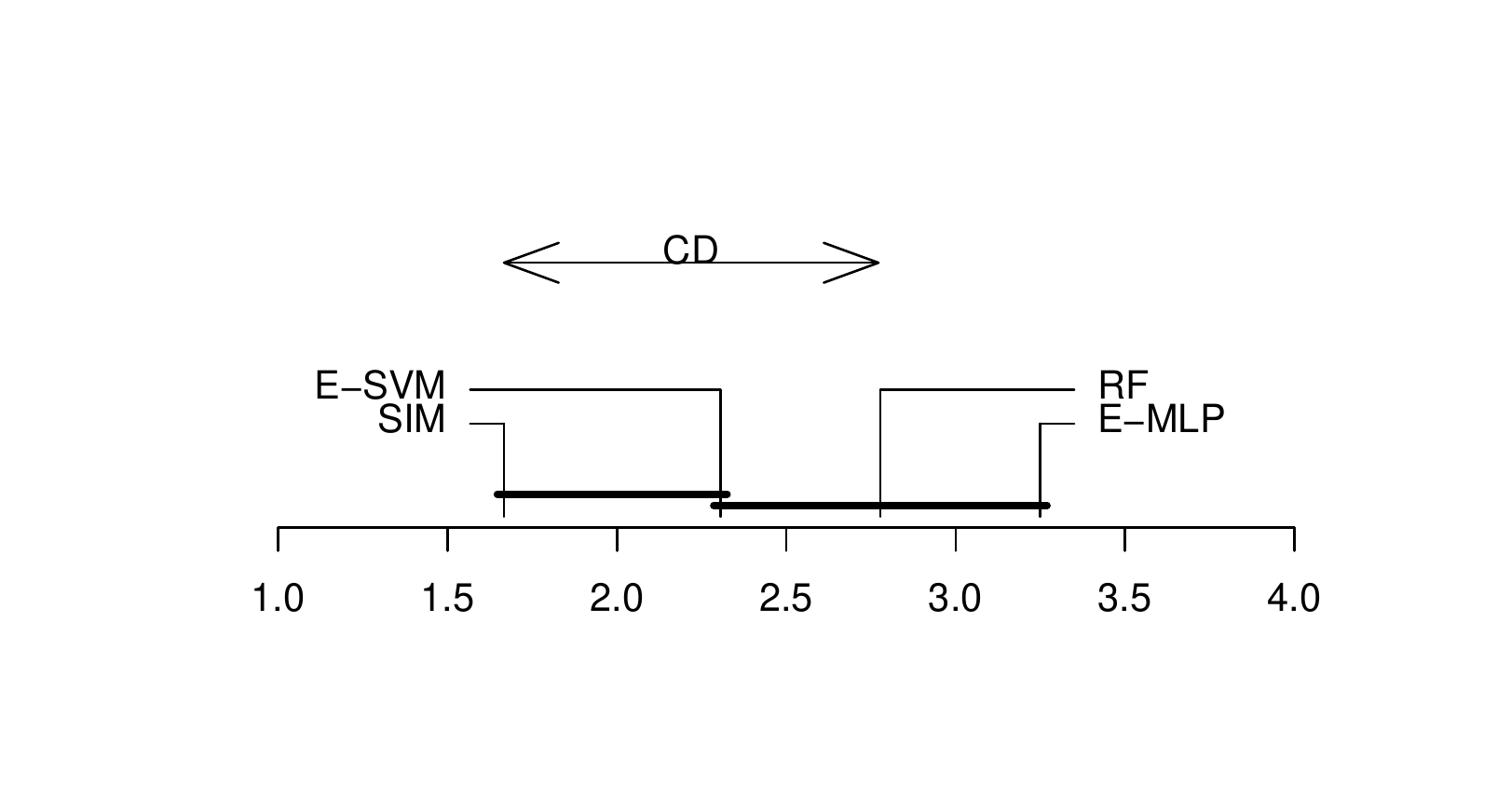}
	\caption{Average ranks for E-SVM, E-MLP, RF and the optimal estimated heterogeneous ensemble  
	}\label{demsar_final}
\end{figure}

As shown in Table~\ref{tab:errors}, the proposed method 
is the best or the second best method for all datasets. E-SVM also achieves
rather good results but it is somehow less consistent. E-SVM is the method that
obtains the highest number of best performances (in 9 datasets) but its
performance is the worst in 4 datasets. Finally, random forest and E-MLP obtain
5 and 1 best results respectively. This results are summarized using a
Dem\v{s}ar plot \cite{demsar_2006_statistical} in Fig.~\ref{demsar_final}.
From this diagram, it can be observed that the proposed procedure is
significantly better than random forest and E-MLP (as given by a Nemenyi test
with p-value $<$ 0.05). The proposed methodology has an average rank better that
E-SVM but their difference is not statistically significant.

\section{Conclusions}		

In this study, a continuous family of heterogeneous ensembles of size T with varying
propositions of base classifiers of different types is analyzed. 
To this end, we first generate
M different homogeneous ensembles. Diversification in these 
ensembles is obtain by using both subsampling and randomization techniques.
Then a heterogenous ensemble is built by pooling classifiers
from these homogeneous ensembles.
The proportions of classifiers of different types in the 
heterogeneous combination can be represented
with a point in a simplex in M dimensions. Each of the M vertices in this simples
corresponds to one of the homogeneous ensembles. 
The optimal proportion of base classifiers 
in the final ensemble, which is strongly problem-dependent,
can be estimated using out-of-bag data. 

In the empirical evaluation carried out, the proposed strategy 
consistently exhibits excellent performance. 
In the problems investigated, it is either the first or second most
accurate method. The results show that the proposed combination is
better that any of the homogeneous ensembles; i.e. random forest, ensembles
of MLPs and ensembles of SVMs. In addition, the differences of average
ranks are statistically significant except for the ensemble of SVMs,
which is second best.

\vspace{12pt}

\bibliographystyle{IEEEtran}
\bibliography{IEEEabrv,ijcnn19}
\end{document}